\documentclass[10pt,conference,english,twocolumn]{IEEEtran} 
\usepackage{babel}
\usepackage[babel]{microtype}
\usepackage[babel]{csquotes}

\usepackage[T1]{fontenc}

\usepackage[svgnames]{xcolor}
\usepackage{graphicx}
\usepackage{tikz}
\usepackage{pgfplots}

\usepackage{tabularx}
\usepackage{booktabs}

\usepackage{amsmath}
\usepackage{amsfonts}
\usepackage{amssymb}
\usepackage{amsthm}
\usepackage{bm}
\usepackage{siunitx}
\usepackage{mathtools}
\usepackage{dsfont}

\usepackage[math]{blindtext}

\usepackage{cite}

\usepackage{hyperref}
\hypersetup{
	pdfauthor={{{author}}},
	pdftitle={{{title}}},
	pdflang={en-US},
	bookmarksnumbered=true,
	colorlinks=true,
	allcolors=.,
	urlcolor=MidnightBlue,
}

\usepackage[acronyms,nomain,xindy,nowarn]{glossaries}
\makeglossaries
\newacronym{5g}{5G}{Fifth-Generation Wireless Systems}
\newacronym{ai}{AI}{Artificial Intelligence}
\newacronym{ann}{ANN}{Artificial Neural Network}
\newacronym{ask}{ASK}{amplitude-shift keying}
\newacronym[plural={AWGN}, \glsshortpluralkey={AWGN}]{awgn}{AWGN}{additive white Gaussian noise}

\newacronym{bec}{BEC}{binary erasure channel}
\newacronym{ber}{BER}{bit error rate}
\newacronym{bler}{BLER}{block error rate}
\newacronym{bpsk}{BPSK}{binary phase-shift keying}
\newacronym{bs}{BS}{base station}

\newacronym{cdf}{CDF}{cumulative distribution function}
\newacronym{cnn}{CNN}{convolutional neural network}
\newacronym{csi}{CSI}{channel state information}
\newacronym{csit}{CSI\babelhyphen{nobreak}T}{channel-state information at the transmitter}
\newacronym{csir}{CSI\babelhyphen{nobreak}R}{channel-state information at the receiver}
\newacronym{comp}{CoMP}{coordinated multi-point}

\newacronym{dnn}{DNN}{deep neural network}
\newacronym{ddpg}{DDPG}{deep deterministic policy gradient}
\newacronym{dqn}{DQN}{deep Q-network}
\newacronym{da2g}{DA2G}{direct air to ground}
\newacronym{ecc}{ECC}{error-correcting code}
\newacronym{ecdf}{ECDF}{empirical distribution function}
\newacronym{ee}{EE}{energy efficiency}

\newacronym{ff}{FF}{feed-forward}
\newacronym{ffnn}{FF-NN}{feed-forward neural network}
\newacronym{fsk}{FSK}{frequency-shift keying}

\newacronym{gm}{GM}{Gaussian mixture}

\newacronym{iid}{i.i.d.}{independent and identically distributed}
\newacronym{il}{IL}{information leakage}
\newacronym{iot}{IoT}{internet of things}

\newacronym{hom}{HOM}{handover margin}

\newacronym{lb}{LB}{lower bound}
\newacronym{los}{LoS}{line-of-sight}

\newacronym{mac}{MAC}{multiple access channel}
\newacronym{map}{MAP}{maximum a posteriori}
\newacronym{mc}{MC}{Monte Carlo}
\newacronym{mimo}{MIMO}{multiple-input multiple-output}
\newacronym{miso}{MISO}{multiple-input single-output}
\newacronym{ml}{ML}{machine learning}
\newacronym{mmwave}{mmWave}{millimeter wave}
\newacronym{mop}{MOP}{multi-objective programming problem}
\newacronym{mrc}{MRC}{maximum ratio combining}
\newacronym{msc}{MSC}{modulation and coding scheme}
\newacronym{mse}{MSE}{mean squared error}
\newacronym{mro}{MRO}{mobility robustness optimization}

\newacronym{nlos}{NLoS}{non-line-of-sight}
\newacronym{nn}{NN}{neural network}
\newacronym{nve}{NVE}{normalized validation error}

\newacronym{pdf}{PDF}{probability density function}
\newacronym{pwc}{PWC}{polar wiretap code}

\newacronym{qam}{QAM}{quadrature amplitude modulation}

\newacronym{ra}{RA}{rearrangement algorithm}
\newacronym{rbf}{RBF}{radial basis function}
\newacronym{relu}{ReLU}{rectified linear unit}
\newacronym{ris}{RIS}{reconfigurable intelligent surface}
\newacronym{rl}{RL}{reinforcement learning}
\newacronym{rnn}{RNN}{recurrent neural network}
\newacronym{rrb}{RRB}{radio resource block}
\newacronym{rsrp}{RSRP}{reference signal received power}
\newacronym{rsrq}{RSRQ}{reference signal received quality}

\newacronym{sac}{SAC}{soft actor-critic}
\newacronym{sgd}{SGD}{stochastic gradient descent}
\newacronym{sgp}{SGP}{secure goodput}
\newacronym{simo}{SIMO}{single-input multiple-output}
\newacronym{sinr}{SINR}{signal-to-interference-plus-noise ratio}
\newacronym{siso}{SISO}{single-input single-output}
\newacronym{snr}{SNR}{signal-to-noise ratio}
\newacronym{svm}{SVM}{support vector machine}
\newacronym{son}{SON}{self-organizing networks}

\newacronym{tvd}{TVD}{total variation distance}
\newacronym{ttt}{TTT}{time to trigger}

\newacronym{uav}{UAV}{unmanned aerial vehicle}
\newacronym{ub}{UB}{upper bound}
\newacronym{urllc}{URLLC}{ultra-reliable low latency communication}

\newacronym{v2v}{V2V}{vehicle-to-vehicle}
\newacronym{v2x}{V2X}{vehicle-to-everything}

\newacronym{zoc}{ZOC}{zero-outage capacity}
\setacronymstyle{long-short}
\glsdisablehyper

\definecolor{plot0}{HTML}{004488}
\definecolor{plot1}{HTML}{DDAA33}
\definecolor{plot2}{HTML}{BB5566}
\definecolor{plot3}{HTML}{000000}
\definecolor{plot4}{HTML}{AAAAAA}

\pgfplotsset{compat=newest}
\pgfplotscreateplotcyclelist{lineplot cycle}{ %
	{plot0, mark=*, thick, mark options=solid},
	{plot1, mark=triangle*, thick, mark options=solid},
	{plot2, mark=square*, thick, mark options=solid},
	{plot3, mark=diamond*, thick, mark options=solid},
	{plot4, mark=pentagon*, thick, mark options=solid},
}

\pgfplotsset{%
	betterplot/.style={
		width=.93\linewidth,
		height=.27\textheight,
		xlabel near ticks,
		ylabel near ticks,
		cycle list name=lineplot cycle,
		mark options=solid,
		xmajorgrids=true,
		xminorgrids=true,
		ymajorgrids=true,
		grid style={line width=.1pt, draw=gray!20},
		major grid style={line width=.25pt,draw=gray!30},
		legend cell align=left,
		legend style = {
			/tikz/every even column/.append style={column sep=0.33cm}
		},
	},
}
\usetikzlibrary{positioning,calc,external}

\newcommand{\todo}[2][]{\ignorespaces\leavevmode
	\if\relax\detokenize{#1}\relax
	{\color{red}[TODO: #2]}%
	\else
	{\color{red}[TODO (#1): #2]}%
	\fi
}

\definecolor{change}{HTML}{0096b8}

\theoremstyle{plain}

\theoremstyle{definition}

\newtheorem*{prob*}{Problem Statement}

\theoremstyle{remark}

\newtheorem*{rem*}{Remark}

\newtheoremstyle{example}{\topsep}{\topsep}{}{}{\itshape}{.}{ }{}
\theoremstyle{example}

\newtheorem*{example*}{Example}

\addto\extrasenglish{%

}

\IfPackageLoadedTF{titlesec}{%
	\ifCLASSOPTIONconference%
	\titlespacing{\section}{0pt}{1.5ex plus 1.5ex minus 0.5ex}{0.7ex plus 1ex minus 0ex} 
	\titlespacing{\subsection}{0pt}{1.5ex plus 1.5ex minus 0.5ex}{0.7ex plus .5ex minus 0ex} 
	\else
	\titlespacing{\section}{0pt}{3.0ex plus 1.5ex minus 1.5ex}{0.7ex plus 1ex minus 0ex} 
	\titlespacing{\subsection}{0pt}{3.5ex plus 1.5ex minus 1.5ex}{0.7ex plus .5ex minus 0ex} 
	\fi
	
	\def\thesubsubsectiondis{\arabic{subsubsection})}
	\def\theparagraphdis{\alph{paragraph})}
	\titleformat{\subsubsection}[runin]{\normalfont\normalsize\itshape}{\thesubsubsectiondis}{.5em}{}[:]
	\titlespacing*{\subsubsection}{\parindent}{0ex plus 0.1ex minus 0.1ex}{1ex}
	\titleformat{\paragraph}[runin]{\normalfont\normalsize\itshape}{\theparagraphdis}{.5em}{}[:]
	\titlespacing*{\paragraph}{2\parindent}{0ex plus 0.1ex minus 0.1ex}{1ex}
}{}

\begin{document}
\title{{{Explainable AI for UAV Mobility Management: A Deep Q-Network Approach for Handover Minimization}}}

	\author{Irshad A. Meer\textsuperscript{*}, Bruno Hörmann\textsuperscript{$\dagger$}, Mustafa Ozger\textsuperscript{$\ddagger$*}, Fabien~Geyer\textsuperscript{\S}, \\ Alberto~Viseras\textsuperscript{$\dagger$}, Dominic~Schupke\textsuperscript{\S}, Cicek Cavdar\textsuperscript{*}
		\\
		\textsuperscript{*}KTH Royal Institute of Technology, Sweden \;\;
  		\textsuperscript{$\dagger$}Motius, Germany
\;\;
  	\textsuperscript{$\ddagger$}Aalborg University, Denmark 
    \\
    \textsuperscript{$\S$}Airbus Central Research and Technology, Germany 
		\\Email: \textsuperscript{*}\{iameer, cavdar\}@kth.se, \textsuperscript{$\dagger$}\{bruno.hoermann, alberto.viseras\}@motius.de, \\  \textsuperscript{$\ddagger$}mozger@es.aau.dk, \textsuperscript{\S}\{fabien.geyer, dominic.schupke\}@airbus.com	}

\maketitle
\thispagestyle{empty}
\begin{abstract}\noindent\boldmath
The integration of unmanned aerial vehicles (UAVs) into cellular networks presents significant mobility management challenges, primarily due to frequent handovers caused by probabilistic line-of-sight conditions with multiple ground base stations (BSs). To tackle these challenges, reinforcement learning (RL)-based methods, particularly deep Q-networks (DQN), have been employed to optimize handover decisions dynamically. However, a major drawback of these learning-based approaches is their black-box nature, which limits interpretability in the decision-making process. This paper introduces an explainable AI (XAI) framework that incorporates Shapley Additive Explanations (SHAP) to provide deeper insights into how various state parameters influence handover decisions in a DQN-based mobility management system. By quantifying the impact of key features such as reference signal received power (RSRP), reference signal received quality (RSRQ), buffer status, and UAV position, our approach enhances the interpretability and reliability of RL-based handover solutions. To validate and compare our framework, we utilize real-world network performance data collected from UAV flight trials. Simulation results show that our method provides intuitive explanations for policy decisions, effectively bridging the gap between AI-driven models and human decision-makers.
\end{abstract}
\glsresetall

\section{Introduction}
The integration of \glspl{uav} into cellular networks introduces unique mobility management challenges. Unlike terrestrial users, \glspl{uav} experience frequent handovers due to their 3D mobility and the fragmented, overlapping coverage of terrestrial \glspl{bs} in the sky~\cite{Stanczak_mobility}. These frequent handovers not only disrupt service continuity but also generate excessive control signaling, degrading overall network efficiency~\cite{3gpp_lte, handover_ch}. 

In contrast to terrestrial users, who primarily consume large amounts of downlink data, UAVs might generate substantial uplink data due to applications such as surveillance ~\cite{baltaci2022analyzing}. 
Traditional handover decision metrics such as \gls{rsrp} and \gls{rsrq} are insufficient, as they can lead to unnecessary handovers and ping-pong effects for the \glspl{uav} due to higher \gls{los} probability with increased number of candidate \glspl{bs}. 
Unnecessary handovers can lead to uplink data loss due to the limited buffer capacity of \glspl{uav}. To mitigate this, integrating the buffer queue status of \glspl{uav} into the handover decision process can improve mobility management~\cite{meer2024mobility}. By considering buffer status, the handover policy can postpone handovers for \glspl{uav} with empty buffers while prioritizing timely transitions to stronger \glspl{bs} when the buffer is full, ensuring faster data transmission.

\Gls{rl} has emerged as a promising approach for optimizing mobility management in wireless networks, as it can learn dynamic policies that balance conflicting objectives, such as minimizing handover rates while ensuring reliable connectivity. Previous studies have demonstrated the effectiveness of \gls{rl}-based techniques in optimizing \gls{uav} handovers. For example, \gls{dqn} has been employed to reduce handovers while preserving signal quality \cite{chen2020efficient, azari2020machine, jang2022proactive}. However, despite their adaptability, \gls{rl} models often function as black boxes, making it difficult to interpret their decision-making process. This lack of transparency presents a significant challenge for practical deployment, especially in safety-critical applications.

The integration of AI in wireless networks, particularly in emerging 6G systems, is inevitable. However, the opacity of AI-driven decision-making raises concerns about trust, validation, and regulatory compliance. Explainable AI (XAI) methods aim to provide insight into the internal reasoning of AI models, improving debugging, performance analysis, and decision validation. Among various XAI techniques, Shapley Additive Explanations (SHAP) offers a mathematically grounded approach to assess feature importance in AI models \cite{guo2020explainable, lundberg2020local}.

While XAI has been applied in wireless networks for tasks such as diagnosing network issues, predicting service-level agreement (SLA) violations, modulation and coding scheme, and optimizing resource allocation~\cite{terra2020explainability, barnard2022resource, barnard2022robust, rezazadeh2023explanation, rjoub2022explainable, xai-mcs}, its application to \gls{uav} mobility management remains largely unexplored. The potential of XAI for feature selection and decision interpretability in the context of \gls{uav} handover decisions is yet to be fully realized.

In this paper, we propose an explainable \gls{rl}-based handover management framework for cellular-connected \glspl{uav}. Unlike conventional approaches, our method integrates buffer state information alongside traditional handover metrics like \gls{rsrp} and \gls{rsrq}, enabling more adaptive and intelligent decision-making. To optimize the handover process, we employ \gls{dqn}, and to enhance transparency, we apply SHAP to quantify the influence of different features on the learned policies. This integration improves both the interpretability and reliability of the model. Additionally, we incorporate language models to generate natural language explanations based on SHAP values, making the \gls{rl} decisions more accessible and comprehensible. We validate our approach using real-world LTE data from \gls{uav} flight trials, demonstrating its effectiveness in improving handover efficiency. By balancing performance and transparency, our framework ensures that \gls{uav} handover decisions are not only optimal but also explainable and trustworthy. Our findings offer valuable insights into AI-driven mobility management, paving the way for more reliable and interpretable \gls{uav} communications in future wireless networks.

\section{System Model and Problem Formulation} 
\label{systemmodel}

\subsection{System Model}

We consider an uplink communication scenario where $K$ \glspl{uav} transmit to $N$ fixed \glspl{bs} over orthogonal resource blocks. Each \gls{uav}~$k$ has a set of candidate handover \glspl{bs}, denoted as $\mathcal{L}_{k}$. The \glspl{bs} operate over a total bandwidth $\mathbb{B}$, which consists of $N_b$~\glspl{rrb}. 

The \gls{da2g} channel model from 3GPP \cite{3gpp_lte} is adopted, which differentiates between \gls{los} and non-\gls{nlos} propagation. The probability of an \gls{los} link between \gls{uav} $k$ and \gls{bs} $l$ is given by:
\begin{equation}
\mathcal{P}_\text{LOS} =
\begin{cases} 
1, & d_{\text{2D},l} \leq d_1 \\
\frac{d_1}{d_{\text{2D},l}} + \exp \left( \frac{d_1}{p_1} \right) \left( 1 - \frac{d_1}{d_{\text{2D},l}} \right), & d_{\text{2D},l} > d_1
\end{cases}
\end{equation} where $d_{\text{2D},l}$ represents the two-dimensional (2D) distance between \gls{uav} $k$ and \gls{bs} $l$, and $p_1$, $d_1$ are altitude-dependent parameters. For \gls{uav} altitude $h > 100$ m, we assume $\mathcal{P}_\text{LOS} = 1$. The corresponding path loss models are:
\begin{small}
\begin{align}
\text{PL}_\text{LOS} &= 28 + 22 \log_{10}(d_{\text{3D},l}) +20 \log_{10}(f_c), \\
\text{PL}_\text{NLoS} &= 15 + (46 - 7\log_{10}(h)) \log_{10}(d_{\text{3D},l}) + 20 \log_{10} (f_c),
\end{align}
\end{small} where $d_{\text{3D},l}$ is the three-dimensional (3D) distance between UAV $k$ and BS $l$, and $f_c$ is the carrier frequency in GHz.
Shadow fading follows $\sigma_\text{LOS} = 4.64 \exp(-0.00066h)$ and $\sigma_\text{NLoS} = 6$. 

Given the above wireless channel, for calculating the received signal power, the \gls{rsrp} at UAV $k$ from BS $l$ is given by:
\begin{equation}
\text{RSRP}_{k,l}(t) = \frac{P_{\text{TX},l} G_k G_l}{\text{PL}_{k,l}},
\end{equation}
where $P_{\text{TX},l}$ is the transmission power of BS $l$, $G_k$ and $G_l$ are the antenna gains of UAV $k$ and BS $l$, respectively, and $\text{PL}_{k,l}$ is the path loss. Correspondingly for the received signal quality, \gls{rsrq} is defined as:
\begin{equation}
\text{RSRQ}_{k,l}(t) = \frac{\text{RSRP}_{k,l}(t)}{I_{k,l}(t) + N_0},
\end{equation} where $I_{k,l}(t)$ represents the total interference power from other BSs using the same frequency:

\begin{equation}
I_{k,l}(t) = \sum_{l' \neq l} P_{\text{TX},l'} G_k G_{l'} \cdot \text{PL}_{k,l'}^{-1}.
\end{equation}

To model the buffer queue status of UAVs, we assume that incoming data packets arrive according to a Poisson point process with a mean arrival rate of $\lambda$. The UAV buffer holds two types of packets: application data packets of size $F_d$ bits and control packets of size $F_c$ bits, where control packets are generated during handover events. The queue evolution follows:
\begin{small}
\begin{equation}
q_k(t+1) = \left[q_k(t)  + \sum_{n=1}^{+\infty} p\left(M=n\vert\lambda\right) \cdot M F_{d} + I(t)\cdot F_{c} - s_k(t)\right]^+,
\end{equation}
\end{small} where $[x]^+ = \max(x,0)$, $p(M=n\vert\lambda) = \frac{e^{- \lambda}\cdot \lambda^{n}}{n!}$ is the probability of generating $M$ packets in a time step, and $I(t) \in \{0,1\}$ indicates a handover event. The transmitted data $s_k(t)$ is given by:

\begin{equation}
s_k(t) = W_s |\mathbf{B}_k(t)| \log_2 (1 + \Gamma_k(t)),
\end{equation} where $W_s$ is the bandwidth per RRB, $\mathbf{B}_k(t)$ is the set of allocated RRBs, and $\Gamma_k(t)$ is the signal-to-interference-plus-noise ratio (SINR) at the serving BS.
The handover decision directly impacts the channel quality to the serving BS, thereby regulating the transmitted data rate \( s_k(t) \), which subsequently influences the buffer data.

\subsection{Problem Formulation}
In this paper, our goal is to minimize the number of handovers and the buffer data of the \gls{uav} while at the same time maintain the service continuity. 
Hence, we define a binary BS association variable, $\Phi_{k,l}(t) \in \{0,1\}$, where $\Phi_{k,l}(t) = 1$ if at time $t$ UAV $k$ is associated with BS $l$, and $0$ otherwise. 
The corresponding optimization problem is defined as follows:
\begin{align}
\label{optimzEq}
&\min_{\Phi_{k,l}(t)} \bigg[ \sum_{k}\sum_{t} q_k(t) + \sum_{k} \sum_{t} I_k(t)\bigg],\ \\
&\textrm{subject to: } \nonumber \\
&(\textbf{C1})\quad  \sum_{l \in \mathcal{L}_k} \Phi_{k,l}(t) = 1, \quad \forall k, t  \nonumber \\
&(\textbf{C2})\quad  q_k(t) \leq q_{\max}, \quad \forall k, t  \nonumber \\
&(\textbf{C3})\quad \text{RSRP}_{k,l}(t) \geq \text{RSRP}_\text{th}, \quad \forall k, l, t \nonumber \\
& (\textbf{C4}) \quad \text{RSRQ}_{k,l}(t) \geq \text{RSRQ}_\text{th}, \quad \forall k, l, t \nonumber
\end{align}
where $q_k(t)$ is the amount of data in the $k$-th \gls{uav} buffer and $I_k(t)$ is an indicator function which is equal to $1$, if the \gls{uav} experiences an handover at time $t$.
(\textbf{C1}) ensures that each UAV is always associated with exactly one BS at any given time. (\textbf{C2}) limits the buffer occupancy of each UAV to prevent excessive queuing delays. (\textbf{C3}) and (\textbf{C4}) enforce that a UAV can only connect to a BS if the received signal power (RSRP) and quality (RSRQ) exceed their respective thresholds, ensuring a reliable communication link.
By satisfying these constraints, the optimization balances minimizing buffer congestion and reducing frequent handovers while maintaining connectivity and service quality for UAVs.
\section{UAV Handover Management with Reinforcement Learning}
\label{sec:dqn_based}
In this section, we provide an \gls{rl}-based solution to solve the optimization problem in \eqref{optimzEq} by 
employing a centralized training and distributed execution (CTDE) framework. 
We assume that a central agent responsible for mobility management is trained by running multiple flights of the \gls{uav} over the serviced area.
Using the trained model, each \gls{uav} is able to take the handover decisions.

We employ \gls{dqn} as the  \gls{rl} algorithm, which optimizes an agent’s behavior through trial-and-error learning. By leveraging past experiences, \gls{dqn} iteratively enhances system performance. In this framework, the agent interacts with the environment over a sequence of events, selecting actions from a predefined set and receiving rewards based on its decisions. The reinforcement learning problem is defined by the tuple $(\mathcal{S}, \mathcal{A}, r)$, where $\mathcal{S}$ represents the set of possible states, $\mathcal{A}$ denotes the available actions, and the agent's objective is to maximize the cumulative reward $r$.
For our specific problem, the \gls{dqn}-based learning framework is defined as follows:\\
\emph{1) The State Space:} 
It describes the set of all the states accessible to the agent upon executing an action within the specified environment. 
The discrete state is characterized by factors influencing handover decisions in a given radio environment. 
In our context, the state of the environment for \gls{uav} $k$ at time $t$ is denoted as $S_k(t)=(x_k(t), y_k(t), z_k(t), q_k(t), SB_k(t), \text{RSRP}_k(t), \text{RSRQ}_k(t))$, encapsulating 
$x$-axis, $y$-axis, and $z$-axis coordinates, buffer queue size, serving BS, RSRP and RSRQ measurements from set of strongest candidate \glspl{bs} of \gls{uav} $k$ respectively.  
To limit the number of states, we quantize the area into small squares, each square center representing the $x$ and $y$ coordinates of a specific state.\\
\emph{2) The Action Space:} The action space presents the set of decision parameters available to the agent at each decision epoch. We present the action at time $t$ for \gls{uav} $k$ as  $a_k(t)$=[$\Phi_{k,l}$], where $\Phi$ stands for \gls{bs} association at time $t$.\\
\emph{3) The Reward Function:} To calculate the  reward~$r_k(t)$ for the \gls{uav} $k$  at time $t$ and make it a positive quantity, we formulate the function as:
\begin{multline}\label{eq:reward}
    r_k(t) =\frac{\omega_1}{1 + q_k(t + 1)} -\omega_2 I(k,t) \\ + \omega_3\sum_{l \in \mathcal{L}_k} \Phi_{k,l}\mathbb{I}(\text{RSRP}_{k,l} \geq \text{RSRP}_\text{th}) \\ + \omega_4\sum_{l \in \mathcal{L}_k} \Phi_{k,l}\mathbb{I}  (\text{RSRQ}_{k,l} \geq \text{RSRQ}_\text{th}),
\end{multline}
where the non-negative weights $\omega_{i}$, $i\in\{1, 2, 3,4 \}$, are used to balance between the individual objectives. 
The first term in reward~$r_k(t)$ in \eqref{eq:reward} 
rewards the agent for reducing the data in the buffer.
The second term penalizes the agent for performing handovers.
The third and forth term rewards the agent for maintaining a reliable connection between the UAV and its associated BS.
Overall, the reward function is designed to encourage the agent to minimize buffer congestion and reduce frequent handovers while maintaining connectivity and service quality for UAVs. \\
\emph{4) Learning Policy:} 
\Gls{rl} enables agents to make sequential decisions by learning a policy that maps states to actions \cite{sutton2018reinforcement}. In this work, we employ a model-free \gls{dqn} approach, where the agent learns an optimal strategy by estimating the state-action value function \( Q_k^\pi(s, a) \), which represents the expected cumulative reward for taking action \( a \) in state \( s \). Instead of maintaining a tabular Q-function, we approximate \( Q(s, a) \) using a neural network.

The DQN consists of two deep neural networks: the main network (\(\theta\)) estimates \( Q(s, a) \), while the target network (\(\theta'\)) stabilizes training by updating its weights every \( M \) iterations. The goal is to maximize the cumulative discounted reward:  
\begin{equation}
    R_{t_0} = \sum_{t=t_0}^{\infty} \gamma^{t - t_0} r_t,
\end{equation}
where \( \gamma \) is the discount factor ensuring future rewards are considered. The action-value function follows Bellman’s equation:  
\begin{equation}
    Q_k(s_k,a_k;\theta) = r + \gamma\max_{\mathbf{a'}\in \mathcal{A}} Q_k(s_k',a_k';\theta').
\end{equation}
Training uses experience replay, where past experiences \((s, a, r, s')\) are stored and randomly sampled to reduce correlation between updates. The loss function is defined as:
\begin{equation}
    L(\theta) = \mathbb{E}[(r + \gamma\max_{\mathbf{a'}\in \mathcal{A}} Q_k(s_k',a_k';\theta') - Q_k(s_k,a_k;\theta))^2].
\end{equation}
Gradient descent minimizes this loss \cite{mh}. The first and final network layers correspond to the dimensions of the state and action spaces.  
To balance exploration and exploitation, we use an \(\epsilon\)-greedy policy, where a random action is selected with probability \( \epsilon \) and the greedy action otherwise:
\begin{equation}
\pi_k(s_k, a_k) =
\begin{cases} 
1-\epsilon, & a_k = \arg\max Q_k(s,a) \\
\epsilon/(|\mathcal{A}|-1), & \text{otherwise}.
\end{cases}
\end{equation}

Initially, \(\epsilon = 1\) to encourage exploration, gradually decaying to \( 0.01 \) to favor exploitation.  
We also apply a decaying learning rate to balance training speed and convergence stability:

\begin{equation}
    \alpha(\Gamma) = \frac{\alpha_0}{1 + \eta \Gamma},
\end{equation} where \(\alpha_0\) is the initial learning rate, \(\Gamma\) is the training episode, and \(\eta\) controls decay speed. A well-tuned learning rate prevents slow convergence or premature sub-optimal solutions.

\section{Explainable Decision-Making Framework}

\begin{figure}[htbp!]
    \centering
     \includegraphics[width=0.99\linewidth]{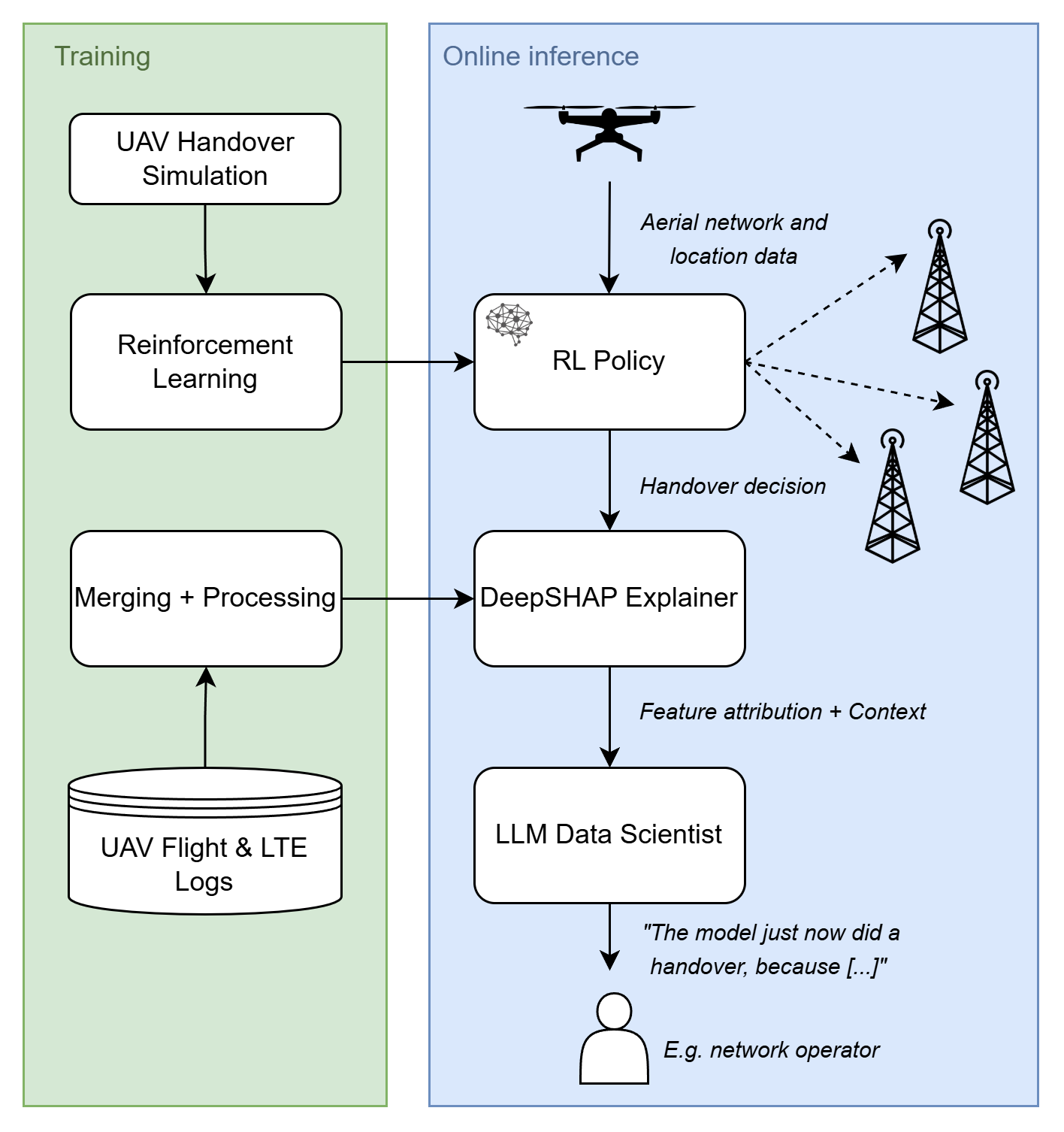}
    \caption{Architecture of the explainable AI (XAI) framework for UAV handover management.}
    \label{fig:xai_framework}
\end{figure}

Although RL-based UAV handover management achieves promising performance, its inherently opaque decision-making process poses significant challenges for trust and regulatory compliance. To address these critical issues, we propose a novel XAI framework, illustrated in Figure~\ref{fig:xai_framework}, composed of three integrated core modules:
\begin{enumerate}
    \item \textbf{Black-Box Model:} A trained \gls{dqn} policy, denoted as \( \pi(s, \theta) \), where \( s \) is the state and \( \theta \) are the deep neural network parameters.
    \item \textbf{SHAP-Based Feature Attribution:} Computation of Shapley values \(\psi_i(\Phi)\) to quantify the contribution of each state parameter \( S_i \) to the BS association decision \( \Phi \).
    \item \textbf{Interpretability Module:} Visualization and natural language synthesis to present explanations to network operators.
\end{enumerate}
This framework improves transparency, accountability, and adherence to regulatory standards in AI-powered mobility management. Our methodology is built upon SHAP \cite{lundberg2017unified}, a well-founded approach that provides a robust theoretical framework for deriving both local and global explanations of machine learning models. 

\vspace{-3pt}
\subsection{Shapley Values for Feature Importance}
Shapley values, rooted in cooperative game theory, provide a fair and mathematically rigorous method to attribute the contribution of each feature to a model's prediction. 
Let $N$ be the set of all features, and let $v(\cdot)$ be the RL model’s prediction which in our case is the BS association decision $\Phi$. The Shapley value $\psi_i$ attributed to feature $i$ is defined as:

\begin{equation}
    \label{eq:shapley}
    \psi_i(v) = \sum_{S \subseteq N \setminus \{i\}} \frac{|S|! \,(|N| - |S| -1)!}{|N|!} \bigl[v\bigl(S \cup \{i\}\bigr) - v(S)\bigr],
\end{equation} where, $S$ represents any subset of features not including $i$, and $v(S)$ is the model's prediction for the feature subset $S$. Intuitively, \eqref{eq:shapley} measures the contribution of feature $i$ by considering all possible subsets of other features.

In practice, exact Shapley value computation can be prohibitively expensive, as it involves evaluating the model on every subset of features. To mitigate this combinatorial explosion, SHAP (SHapley Additive exPlanations) employs various approximations~\cite{lundberg2017unified}. SHAP thus yields both local and global explanations: at the local level, it quantifies feature contributions for individual decisions; at the global level, it aggregates these insights to present an overall feature-importance ranking.

For deep neural networks (as used in a DQN), DeepSHAP combines SHAP with DeepLIFT to propagate contribution scores backward through the network~\cite{shrikumar2017learning}. This technique approximates Shapley values efficiently by leveraging the network’s structure, negating the need for exhaustive evaluations. In this paper, we adopt DeepSHAP for interpreting the learned RL policy. 
We train the DeepSHAP on background data sampled from the simulation environment, serving as a baseline for feature contributions. For a given state \( s \), DeepSHAP quantifies the contribution of each feature \( s_i \) to the logit of each action \( a \), enabling interpretable decision analysis. For instance, in UAV handover scenarios, DeepSHAP can attribute the decision to factors such as signal strength (RSRP/RSRQ), buffer status, altitude, and UAV location.

\subsection{Bridging the Gap with Language Models}

While SHAP offers precise feature attributions, its outputs remain opaque. To improve interpretability, we incorporate language models (LMs) to generate natural language explanations for RL decisions. The LM processes:
\begin{itemize}
    \item Model-predicted action logit values \( \mathbf{z} = f(s, \theta) \).
    \item Input feature values \( \mathbf{x} \).
    \item SHAP-derived feature contributions \( \boldsymbol{\psi} \).
    \item Contextual metadata (e.g., UAV ID, timestamp).
\end{itemize}

Using this structured input, the LM generates human-readable explanations, such as:
\begin{quote}
\textit{``The model initiated a handover due to a 40\% higher RSRP at the target \gls{bs}. Additionally, the empty buffer minimized disruption to ongoing data transmission.''}
\end{quote}
However, LMs introduce risks such as hallucinations, confirmation bias, and cybersecurity vulnerabilities. To mitigate these, we employ:
\begin{itemize}
    \item \textbf{Predefined Explanation Templates:} Constraining LM creativity to ensure factual accuracy.
    \item \textbf{Confidence Scores:} Quantifying the reliability of generated explanations.
    \item \textbf{Human-in-the-Loop Validation:} Detecting and correcting inaccuracies through operator feedback.
\end{itemize}

\begin{figure*}
    \centering
    \includegraphics[width=\linewidth]{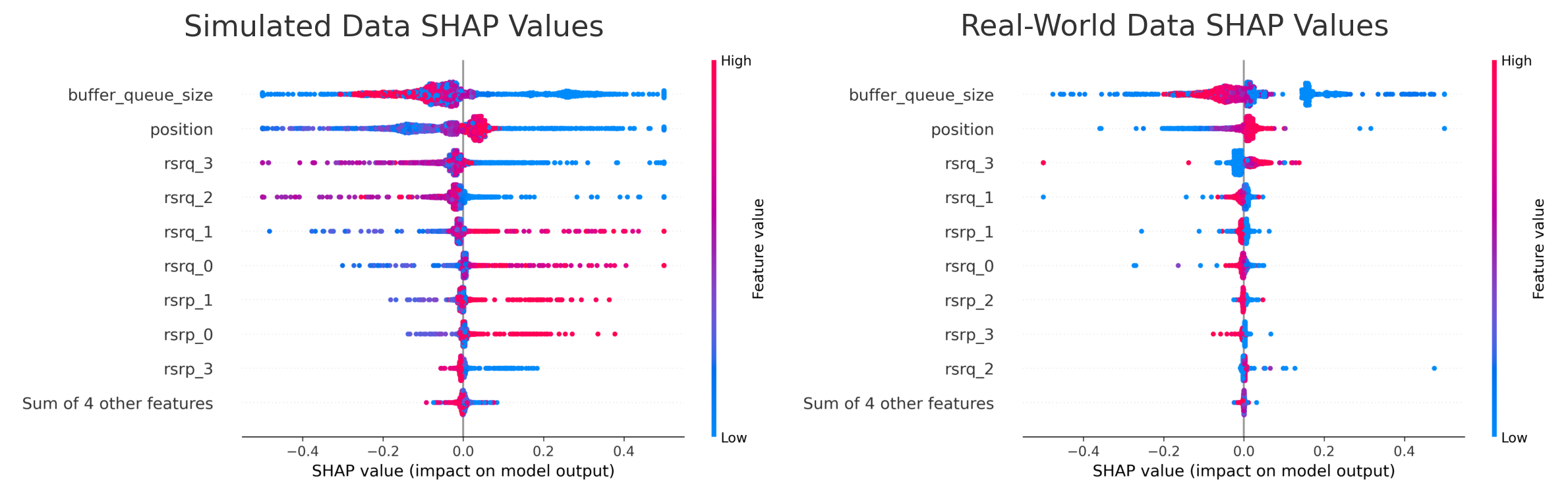}
    \caption{Comparison of SHAP value distributions for simulated (left) and real-world (right) data. Each point represents a data instance, colored by feature value magnitude, illustrating similarities and differences in feature contributions to model outputs across simulation and real-world deployments.}
    \label{fig:shap_comparison}
     \vspace{-1.65mm}
\end{figure*}


\subsection{Evaluation with Real-World Data}

To validate our framework, we evaluated it using real-world LTE network performance data collected during UAV flight trials \cite{baltaci2022analyzing}. The dataset included uplink channel measurements (RSRP, RSRQ) and network parameters recorded under operational conditions with a commercial UAV (DJI-M600). The UAV was equipped with two LTE modems, enabling it to switch between network providers based on channel conditions. The dataset comprises approximately 90 flights recorded in urban and rural settings.


\section{Results}


\vspace{-0.01cm}
\begin{table}[b!]
 \centering
 \caption{Median Performance Across Different Methods}
 \label{tab:handover_results}
\begin{tabular}{c|c c}
\toprule
 \textbf{Methods} 
& {\textbf{Number of handover}} 
& {\textbf{Ping Pong Percentage}}\\
\midrule
\texttt{CHM} & 75  & 40\%  \\
\texttt{SA-MRO} \cite{meer2024mobility}  & 32 &  27\% \\
\texttt{DQN-Based}  & 12 & 14\%  \\

\bottomrule
\end{tabular}
\end{table}

We compare the proposed \texttt{DQN-Based} handover optimization method, detailed in \autoref{sec:dqn_based}, with two baseline approaches: the conventional handover management (\texttt{CHM}) scheme that uses fixed handover parameters, and the model-based Service Availability-based Mobility Robustness Optimization (\texttt{SA-MRO}) approach from \cite{meer2024mobility}. The results, summarizing the number of handovers and the percentage of ping-pong events experienced by the UAV under each scheme, are presented in \autoref{tab:handover_results}. As seen in the table, the \texttt{DQN-Based} approach achieves a reduction of more than $80\%$ in the number of handovers and limits the ping-pong rate to approximately $14\%$. These results demonstrate that the \texttt{DQN-Based} solution successfully learns an effective handover policy for UAVs. However, the primary focus of this work is to interpret the reasoning behind the learned policy. To this end, we analyze the feature importance using SHAP values, computed for both the simulated training data and real-world data. Figure~\ref{fig:shap_comparison} illustrates this comparison, highlighting the influence of each feature on the model's output. 
In this figure, each point represents the SHAP value of a feature for a specific instance. The x-axis reflects the impact of the feature on the model’s prediction, while the color indicates the actual feature value (blue for low, red for high).

In the simulated data (left panel), \verb|buffer_queue_size| emerges as the most influential feature, with high values consistently leading to positive SHAP values. \verb|Position| of the UAV ranks second, showing a similar positive correlation. Signal quality features like \verb|rsrq_3| to \verb|rsrq_0|, representing RSRQ values from candidate BSs 3 to 0, have a more scattered impact, though high \verb|rsrq_3| values tend to contribute positively. In contrast,  the RSRP features (\verb|rsrp_1|, \verb|rsrp_0|, \verb|rsrp_3|) generally exhibit lower overall importance in the simulated dataset compared to \verb|buffer_queue_size| and \verb|position|.

In the real-world data (right panel), \verb|buffer_queue_size| remains the most dominant feature, with an even clearer separation between the effects of high and low values. High values strongly increase predictions, while low values drive them down. \verb|Position| retains its second-place ranking, again with higher values linked to positive SHAP impacts.

The relative importance of RSRQ and RSRP features differs from the simulated data. For instance, \verb|rsrq_3| and \verb|rsrq_1| appear more significant in the real-world case. Moreover, the density of high-impact points for signal quality features is lower in real-world data. An interesting deviation is observed for \verb|rsrp_1|, where low values (blue points) tend to push predictions upward, contrary to the top features.


A comparative analysis of the SHAP plots reveals that while core features like \verb|buffer_queue_size| and \verb|position| consistently exhibit high importance across both datasets with similar directional impacts, the influence of signal quality metrics varies. This discrepancy underscores the differences between the simulated environment and the complexities inherent in real-world conditions. The consistent importance of \verb|buffer_queue_size| and \verb|position| suggests their fundamental role in the underlying process being modeled. However, the shifts in ranking and impact patterns for RSRQ and RSRP features between simulated and real-world data highlight the challenges in perfectly replicating real-world signal dynamics in simulations. 

\begin{figure}
    \centering
    \includegraphics[width=\linewidth]{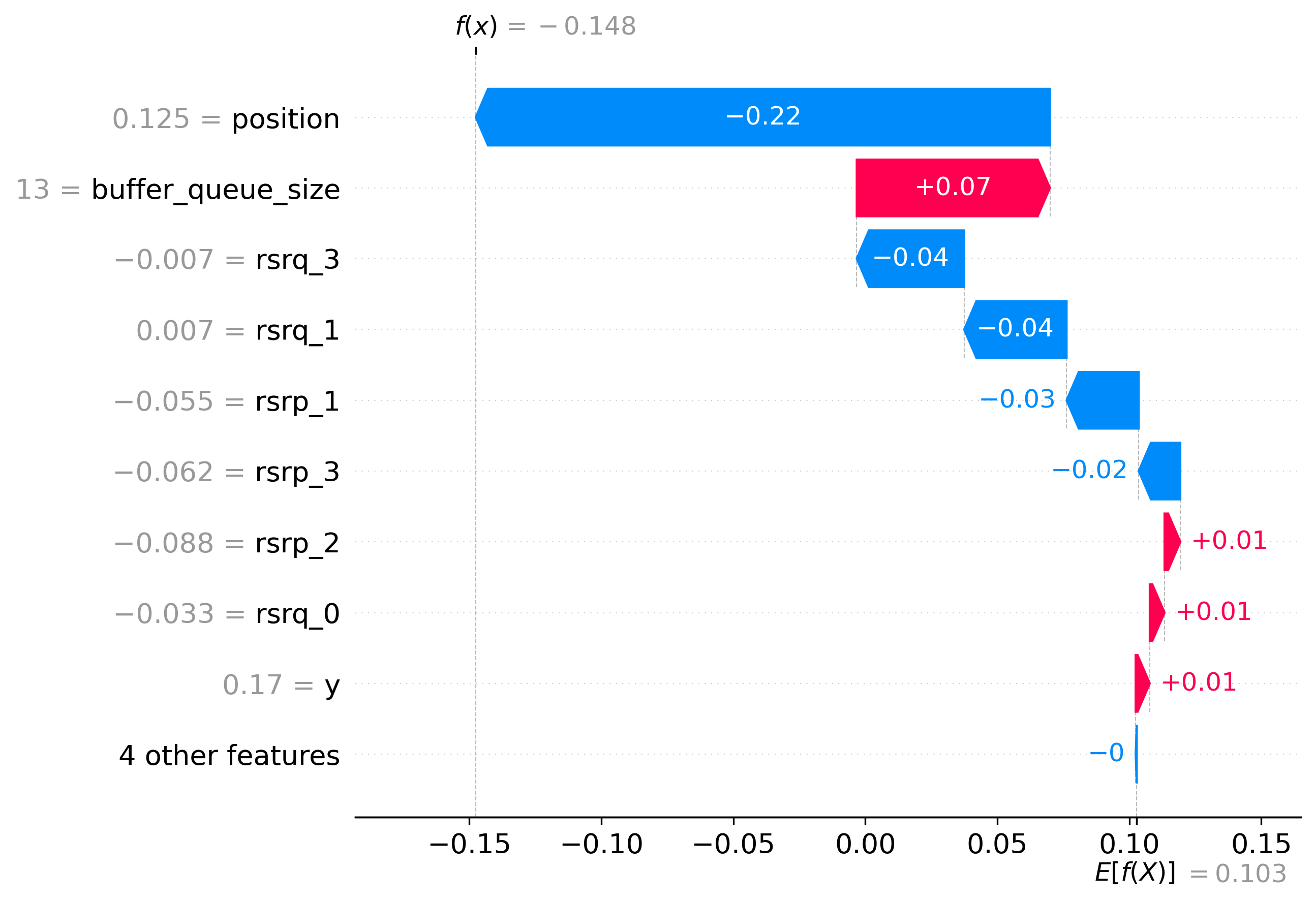}
    \caption{Waterfall plot depicting SHAP values for simulated data, illustrating the contributions of various features to the prediction of an action.}
    \label{fig:shap_value_waterfall}
     \vspace{-1.65mm}
\end{figure}

\autoref{fig:shap_value_waterfall} further illustrates the interpretability of our model by showcasing a SHAP waterfall plot, focusing on the feature contributions for a specific prediction instance. Similar to the SHAP value plot in \autoref{fig:shap_comparison} which revealed varying degrees of feature influence based on distribution width, this instance-level view demonstrates the directional impact and magnitude of individual features on a single output. As depicted, feature \verb|position| exerts the most substantial influence, driving the prediction downwards, while \verb|buffer_queue_size| and other features like \verb|rsrp_0| exert smaller, either positive or negative, influences. This detailed breakdown allows us to understand not only the overall importance of features as shown in \autoref{fig:shap_comparison}, but also their specific roles in shaping individual predictions, highlighting the instance-specific dynamics captured by the model.

\section{Conclusion}
In this paper, we proposed an explainable AI framework for UAV mobility management using a deep Q-network (DQN) approach. By incorporating Shapley Additive Explanations (SHAP), we provided insights into the decision-making process of the reinforcement learning-based handover strategy. Our results demonstrate that integrating buffer status with traditional handover metrics reduces unnecessary handovers while maintaining connectivity. Furthermore, the explainability of our model enhances trust and transparency, making it more suitable for real-world deployment. Future work will focus on extending this framework to multi-agent reinforcement learning for coordinated UAV mobility management.

\section*{Acknowledgment}
This work was partially funded by EU Celtic Next Project, 6G for Connected Sky (6G-SKY) with support of the Federal Ministry for Economic Affairs and Climate Action under the contract number 01MJ22010B, and Vinnova, Swedish Innovation Agency. The views expressed herein can in no way be taken to reflect the official opinion of the German ministry.

\bibliographystyle{IEEEtran}
\bibliography{main}

\begin{thebibliography}{10}
\providecommand{\url}[1]{#1}
\csname url@samestyle\endcsname
\providecommand{\newblock}{\relax}
\providecommand{\bibinfo}[2]{#2}
\providecommand{\BIBentrySTDinterwordspacing}{\spaceskip=0pt\relax}
\providecommand{\BIBentryALTinterwordstretchfactor}{4}
\providecommand{\BIBentryALTinterwordspacing}{\spaceskip=\fontdimen2\font plus
\BIBentryALTinterwordstretchfactor\fontdimen3\font minus \fontdimen4\font\relax}
\providecommand{\BIBforeignlanguage}[2]{{%
\expandafter\ifx\csname l@#1\endcsname\relax
\typeout{** WARNING: IEEEtran.bst: No hyphenation pattern has been}%
\typeout{** loaded for the language `#1'. Using the pattern for}%
\typeout{** the default language instead.}%
\else
\language=\csname l@#1\endcsname
\fi
#2}}
\providecommand{\BIBdecl}{\relax}
\BIBdecl

\bibitem{Stanczak_mobility}
J.~{Stanczak}, I.~Z. {Kovacs}, D.~{Koziol}, J.~{Wigard}, R.~{Amorim}, and H.~{Nguyen}, ``{Mobility Challenges for Unmanned Aerial Vehicles Connected to Cellular LTE Networks},'' in \emph{IEEE VTC Spring}, 2018, pp. 1--5.

\bibitem{3gpp_lte}
\BIBentryALTinterwordspacing
{3GPP TR 36.777}, ``{Enhanced LTE support for aerial vehicles},'' Tech. Rep., 2018. [Online]. Available: \url{ftp://www.3gpp.org}
\BIBentrySTDinterwordspacing

\bibitem{handover_ch}
A.~Fakhreddine \emph{et~al.}, ``Handover challenges for cellular-connected drones,'' in \emph{{ACM 5th Workshop on Micro Aerial Vehicle Networks, Systems, and Applications}}, 2019.

\bibitem{baltaci2022analyzing}
A.~Baltaci, H.~Cech, N.~Mohan, F.~Geyer, V.~Bajpai, J.~Ott, and D.~Schupke, ``Analyzing real-time video delivery over cellular networks for remote piloting aerial vehicles,'' in \emph{ACM Internet Measurement Conference}, 2022, pp. 98--112.

\bibitem{meer2024mobility}
I.~A. Meer, M.~Ozger, D.~A. Schupke, and C.~Cavdar, ``Mobility management for cellular-connected uavs: Model-based versus learning-based approaches for service availability,'' \emph{IEEE Transactions on Network and Service Management}, vol.~21, no.~2, pp. 2125--2139, 2024.

\bibitem{chen2020efficient}
Y.~Chen, X.~Lin, T.~Khan, and M.~Mozaffari, ``{Efficient drone mobility support using reinforcement learning},'' in \emph{IEEE WCNC}, 2020, pp. 1--6.

\bibitem{azari2020machine}
A.~Azari, F.~Ghavimi, M.~Ozger, R.~Jantti, and C.~Cavdar, ``{Machine Learning assisted Handover and Resource Management for Cellular Connected Drones},'' in \emph{IEEE VTC-Spring}, 2020.

\bibitem{jang2022proactive}
Y.~Jang, S.~M. Raza, M.~Kim, and H.~Choo, ``{Proactive handover decision for UAVs with deep reinforcement learning},'' \emph{Sensors}, vol.~22, no.~3, p. 1200, 2022.

\bibitem{guo2020explainable}
W.~Guo, ``Explainable artificial intelligence for 6g: Improving trust between human and machine,'' \emph{IEEE Communications Magazine}, vol.~58, no.~6, pp. 39--45, 2020.

\bibitem{lundberg2020local}
S.~M. Lundberg \emph{et~al.}, ``From local explanations to global understanding with explainable ai for trees,'' \emph{Nature machine intelligence}, vol.~2, no.~1, pp. 56--67, 2020.

\bibitem{terra2020explainability}
A.~Terra, R.~Inam, S.~Baskaran, P.~Batista, I.~Burdick, and E.~Fersman, ``Explainability methods for identifying root-cause of sla violation prediction in 5g network,'' in \emph{IEEE Global Communications Conference}, 2020, pp. 1--7.

\bibitem{barnard2022resource}
P.~Barnard \emph{et~al.}, ``{Resource reservation in sliced networks: An explainable artificial intelligence (XAI) approach},'' in \emph{IEEE ICC}, 2022, pp. 1530--1535.

\bibitem{barnard2022robust}
------, ``{Robust network intrusion detection through explainable artificial intelligence (XAI)},'' \emph{IEEE Networking Letters}, vol.~4, no.~3, pp. 167--171, 2022.

\bibitem{rezazadeh2023explanation}
F.~Rezazadeh \emph{et~al.}, ``Explanation-guided deep reinforcement learning for trustworthy 6{G} {RAN} slicing,'' in \emph{IEEE ICC Workshops}, 2023, pp. 1026--1031.

\bibitem{rjoub2022explainable}
G.~Rjoub, J.~Bentahar, and O.~A. Wahab, ``Explainable ai-based federated deep reinforcement learning for trusted autonomous driving,'' in \emph{IEEE IWCMC}, 2022, pp. 318--323.

\bibitem{xai-mcs}
K.~Hao, J.~M. Cortes, and M.~Ozger, ``{Counterfactual and Causal Analysis for AI-Based Modulation and Coding Scheme Selection},'' in \emph{IEEE GLOBECOM Workshops}, 2023, pp. 32--37.

\bibitem{sutton2018reinforcement}
R.~S. Sutton and A.~G. Barto, \emph{Reinforcement learning: An introduction}.\hskip 1em plus 0.5em minus 0.4em\relax MIT press, 2018.

\bibitem{mh}
V.~Mnih \emph{et~al.}, ``Human-level control through deep reinforcement learning,'' \emph{Nature}, vol. 518, no. 7540, p. 529, 2015.

\bibitem{lundberg2017unified}
S.~M. Lundberg and S.-I. Lee, ``A unified approach to interpreting model predictions,'' \emph{Advances in neural information processing systems}, vol.~30, 2017.

\bibitem{shrikumar2017learning}
A.~Shrikumar, P.~Greenside, and A.~Kundaje, ``Learning important features through propagating activation differences,'' in \emph{International conference on machine learning}.\hskip 1em plus 0.5em minus 0.4em\relax PMlR, 2017, pp. 3145--3153.

\end{thebibliography}
\end{document}